\documentclass[journal]{IEEEtran}  
\let\labelindent\relax         

\usepackage{times}
\usepackage{graphicx}\graphicspath{{fig/}} 
\usepackage{epstopdf}
\usepackage{algorithm}
\usepackage{algorithmic}
\usepackage{multirow} 
\usepackage{multicol}
\usepackage{array}
\usepackage{mdwmath}
\usepackage{mdwtab}
\usepackage{rotating}
\usepackage{caption}
\usepackage{amssymb}
\usepackage{verbatim}
\usepackage{isomath}
\usepackage{overpic}
\usepackage[list=off]{caption}

\usepackage{booktabs} 

\usepackage{siunitx} 
\usepackage{pgfplotstable} 

\usepackage{tikz} 
\usetikzlibrary{shapes,calc,patterns,
	decorations.pathmorphing,
	decorations.markings}

\usepackage[siunitx]{circuitikz} 

\makeatletter\newcommand{\manuallabel}[2]{\def\@currentlabel{#2}\label{#1}}\makeatother

\usepackage{color}
\usepackage{xcolor}

\colorlet   {lightorange}{orange!20}
\colorlet   {lightgrey}  {gray!20}

\usepackage{xspace}
\newcommand{\newabbreviation}[2]{#1\ (\renewcommand{#1}{#2}#1)}

\newcommand{\VIA    }{variable impedance actuator}

\newcommand{\VSA    }{variable stiffness actuator}
\newcommand{\SEA    }{series elastic actuator}

\newcommand{\VIAs   }  {\VIA s}
\newcommand{\MACCEPA}  {Mechanically Adjustable Compliance and Controllable Equilibrium Position Actuator}
\newcommand{\MACCEPAVD}{\MACCEPA\ with Variable Damping}
\newcommand{\ILQR   }  {Iterative Linear Quadratic Regulator}
\newcommand{\VSAs}     {\VSA s}

\usepackage{amssymb}
\usepackage{mathtools}

\mathchardef\mhyphen="2D   

\newcommand{\intd}  {\mathrm{d}}          

\newcommand{\qdot}   {\dot{q}}            

\renewcommand{\nu}  {q}                   

\newcommand  {\dt}         {\,\intd t}             

\newcommand{\inertia}{m} 

\newcommand{\Es}{E_s}

\newcommand{\powerin}{P_{\mathrm{in}}}
\newcommand{\powerout}{P_{\mathrm{out}}}
\newcommand{\powerinone}{P_{\mathrm{in1}}}
\newcommand{\powerintwo}{P_{\mathrm{in2}}}
\newcommand{\Prege}{P_\mathrm{rege}}
\newcommand{\Erege}{E_\mathrm{rege}}
\newcommand{\Ein}{E_\mathrm{in}}
\newcommand{\Enet}{E_\mathrm{net}}

\newcommand{\DCr}{D_r}
\newcommand{\DCd}{D_d}

\newcommand{\motorone}{\mathrm{M}_1}
\newcommand{\motortwo}{\mathrm{M}_2}

\graphicspath{
{figures/}
}
\providecommand{\figurename}{Fig.}
\newcommand*{\sref}[1]{\S\ref{s:#1}}            
\newcommand*{\tref}[1]{\tablename~\ref{t:#1}}   
\newcommand*{\fref}[1]{\figurename~\ref{f:#1}}  
\newcommand*{\eref}[1]{(\ref{e:#1})}            

\usepackage[inline]{enumitem}
\setlist{nolistsep}
\newcommand{\il}[1]{\begin{enumerate*}[label=(\roman*)]#1\end{enumerate*}}

\newcommand{\eg}{\textit{e.g.,}~} %
\newcommand{\ie}{\textit{i.e.,}~} %
\newcommand{\cf}{\textit{cf.}~} %
\newcommand{\etc}{\textit{etc.}}  %
\newcommand{\etal}{\textit{et al.}~} %

\newcommand*{\schemeref}[1]{Scheme \ref{scheme:#1}} 
\usepackage[normalem]{ulem}                                        
\usepackage{marginnote}
\usepackage[textwidth=10ex,colorinlistoftodos]{todonotes}

\colorlet{jb}{red}
\colorlet{mh}{red}
\colorlet{fw}{green}

\providecommand{\edit}[2]{\sout{#1}~{\color{change}#2}}
\renewcommand{\sout}[1]{}
\colorlet{change}{blue}      

\tikzstyle{spring}=[very thick,decorate,decoration={zigzag,pre length=2,post
	length=2,segment length=6}]

\tikzstyle{damper}=[thick,decoration={markings, 
	mark connection node=dmp,
	mark=at position 0.5 with 
	{
		\node (dmp) [thick,inner sep=0pt,transform shape,rotate=-90,minimum
		width=15pt,minimum height=3pt,draw=none] {};
		\draw [thick] ( $(dmp.north east)+(2pt,0)$ ) -- (dmp.south east) -- (dmp.south
		west) -- ( $(dmp.north west)+(2pt,0)$ );
		\draw [thick] ( $(dmp.north)+(0,-5pt)$ ) -- ( $(dmp.north)+(0,5pt)$ );
	}
}, decorate]

\usepackage{amsfonts}
\usepackage{amsmath}
\usepackage{amssymb}
\usepackage{graphicx}
\usepackage{caption}
\usepackage{subcaption}
\usepackage{standalone}
\usepackage{setspace}
\IEEEoverridecommandlockouts

\renewcommand{\baselinestretch}{0.980}
\begin{document}

\title{Energy regenerative damping in variable impedance actuators for long-term robotic deployment}

\author{Fan Wu and Matthew Howard$^{*\dagger}$ 
	\thanks{$^{*}$Fan Wu and Matthew Howard are with the Centre for Robotics Research, Department of Informatics, King's College London, UK {\tt\small \{fan.wu, matthew.j.howard\}@kcl.ac.uk.}}%
	\thanks{$^{\dagger}$This work was supported in part by the UK Engineering and Physical Sciences Research Council (EPSRC) SoftSkills project, EP/P010202/1.}%
}

\markboth{XXXX, ~Vol. X, No. X, January~2020}%
{Shell \MakeLowercase{\textit{et al.}}: Bare Demo of IEEEtran.cls for IEEE Journals}

\maketitle
\IEEEpeerreviewmaketitle
\begin{abstract}
	Energy efficiency is a crucial issue towards long-term deployment of compliant robots in the real world. In the context of \newabbreviation{\VIAs}{VIAs}, one of the main focuses has been on improving energy efficiency through reduction of energy consumption. However, the \emph{harvesting} of dissipated energy in such systems remains under-explored. This study proposes a variable damping module design enabling energy regeneration in \VIAs\ by exploiting the regenerative braking effect of DC motors. The proposed damping module uses four switches to combine \emph{regenerative} and \emph{dynamic} braking, in a hybrid approach that enables energy regeneration without a reduction in the range of damping achievable. A physical implementation on a simple VIA mechanism is presented in which the regenerative properties of the proposed module are characterised and compared against theoretical predictions. To investigate the role of variable regenerative damping in terms of energy efficiency of long-term operation, experiments are reported in which the VIA, equipped with the proposed damping module, performs sequential reaching to a series of stochastic targets. The results indicate that the combination of variable stiffness \emph{and} variable regenerative damping 
	results in a $25\%$ performance improvement on metrics incorporating reaching accuracy, settling time, energy consumption and regeneration over comparable schemes where either stiffness or damping are fixed.
\end{abstract}

\section{Introduction}\label{s:introduction}\noindent
The mass deployment of robotic solutions in manufacturing causes huge energy demand. For example, $8\%$ of the total electrical energy usage in production processes of automotive industries is consumed by industrial robots \cite{Paryanto2015}.
For ecological and economic reasons, this motivates research on reducing the energy cost of industrial robots. Furthermore, with the extensive deployment of compliant robots expected in the near-future for human-robot collaboration, medical and civil services, \etc, this imperative  to save energy is likely to become even more critical. 
Variable impedance actuators (\renewcommand{\VIA}{VIA\xspace}\renewcommand{\VIAs}{VIAs\xspace}\VIAs) are believed to be the key for the next generation of robots to interact safely with uncertain environments and provide better performance in cyclic tasks and dynamical movements \cite{Vanderborght2012}. For example, the physical compliance incorporated in \emph{\VSAs} (\renewcommand{\VSA}{VSA\xspace}\renewcommand{\VSAs}{VSAs\xspace}\VSAs) (\eg using elastic components such as springs) enables energy storage, which can be used to \il{\item absorb external energy introduced into the system (\eg from collisions) to enhance safety, and \item amplify output power by releasing stored energy as and when required by the task \cite{Grebenstein2011}}. 

Recently, much research effort has gone into the design of \emph{variable physical damping} actuation, based on different principles of damping force generation (see \cite{Vanderborght2013,Tagliamonte2012} for a review). 
Variable physical damping has proven to be necessary to achieve better task performance, for example, in eliminating undesired oscillations caused by the elastic elements of \VSAs\ \cite{Laffranchi2012a, Laffranchi2013a}. It has also been demonstrated that variable physical damping plays an important role in terms of energy efficiency for actuators that are required to operate at different frequencies, to optimally exploit the natural dynamics \cite{Laffranchi2012a}. However, while these studies represent important advances in terms of improving the efficiency of \emph{energy consumption} in \VIAs, the importance of variable physical damping may be underestimated, because the potential to \emph{harvest energy dissipated by damping} has so far received little attention.

To address this, the authors' prior work \cite{Wu2018} proposed to extend the variable damping technique introduced by \cite{Radulescu2012} to take into account the energy regeneration capabilities of DC motors. 
In particular, \cite{Wu2018} introduced a circuit design 
that enables adjustment of the electrical damping effect, while increasing the damping range available to the controller. 
A non-monotonic relation of the damping effect and the power of regeneration of the proposed damping module emerges that requires balancing a trade-off between damping allocation and energy regeneration in a non-trivial way. 

This paper significantly extends the work in \cite{Wu2018} by implementing the proposed regenerative damping module on a physical robot driven by a \VIA, to gain deeper insight into the role of variable regenerative damping and investigate the energy efficiency problem in the context of long-term deployment and operation of compliant robots. In contrast to the cyclic movement tasks commonly explored in prior work, this paper presents experiments in performing a stochastic movement task that mimics long term industrial operation and measures the performance of \VIAs designed for versatile purposes.
The results 
demonstrate that variable regenerative damping, \emph{in combination with an optimally exploited variable stiffness mechanism}, can contribute both enhanced dynamic performance and improved energy efficiency (in terms of both consumption and regeneration).
Measuring performance through four metrics (accuracy, settling time, energy consumption and regeneration), results reported here indicate that this approach 
can outperform schemes where stiffness and/or damping are fixed by up to $25\%$.


\section{Background}\label{s:background}\noindent
The problem of energy efficiency in compliant robotic systems has been addressed via different approaches from the perspectives of control or design, which can be mainly categorised into studies that \il{\item look at \emph{exploiting energy storage} in periodic or discrete movements, or \item focus on \emph{reducing energy consumption} via the mechanical design}. In the following subsections, the basic concept of power flow of \VIAs\ is introduced as a means to understand these approaches, and point out a third possible way to improve the energy efficiency by regenerating energy dissipated by damping. This is followed by an account of the theory of regenerative braking of electric motors. 
	\edit{Against this backdrop, a summary of related work on the topic of energy efficiency for such systems is provided in \sref{related_work}.}{}

\subsection{Power flow of \VIAs}\label{s:power-flow}
\begin{figure}[t!]
	\centering
	\begin{overpic}[width=0.68\linewidth]{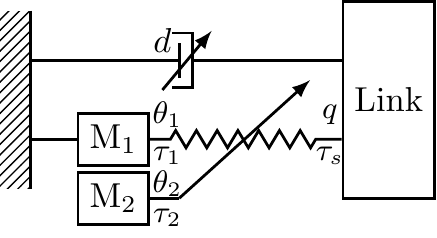}\put(60,0){\ref{f:mass-spring-damper}}\end{overpic} \hfill
	\begin{overpic}[width=0.28\linewidth]{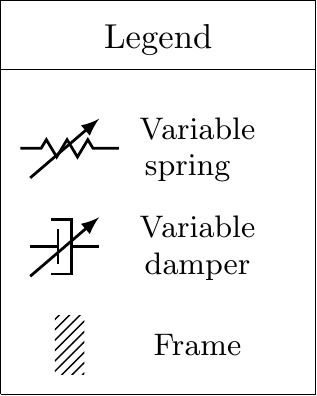}\end{overpic} \hfill%
	\vspace{0.4cm}
	\begin{overpic}[width=0.9\linewidth]{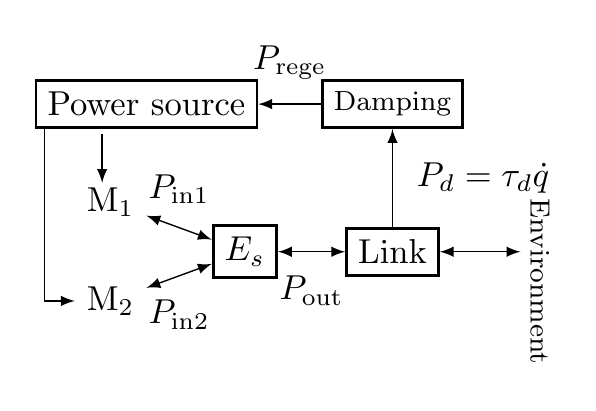}\put(40,0){\ref{f:power-flow}}\end{overpic}%
	\caption{\label{f:model_and_powerflow}Schematic diagrams of 
		\begin{enumerate*}[label=(\alph*)]
			\item\label{f:mass-spring-damper} a \VIA\ represented by a mass-spring-damper model, and
			\item\label{f:power-flow} the corresponding power flow.
		\end{enumerate*} The arrow between damping module and power source shows that the energy dissipated via damping can be harvested and used to recharge the power source.
	}
\end{figure}

\noindent
A large number of mechanical designs have been proposed in the literature to achieve variable impedance actuation \cite{Tagliamonte2012}. This paper focuses on \VIA\ mechanisms that can be represented by the mass-spring-damper model depicted in \fref{model_and_powerflow}\ref{f:mass-spring-damper}. The link, whose position is denoted by $q$, is connected in series to a motor and a variable spring. The equilibrium position is controlled by the motor $\motorone$ and the stiffness is modulated by another motor $\motortwo$, whose positions are denoted by $\theta_1$ and $\theta_2$, respectively. The damper in this system is arranged between the link and the base, and the damping $d$ is independently controllable. The torques exerted by the spring on the link, $\motorone$, and $\motortwo$ are $\tau_s, \tau_{1}, \tau_{2}$.

The corresponding power flow of the system is shown in \fref{model_and_powerflow}\ref{f:power-flow}. A power source is assumed to supply the motors $\mathrm{M}_1,\mathrm{M}_2$. The elastic element, (\ie spring) can be viewed as an energy tank in the actuator that stores potential energy $\Es$. 
\sout{Regulating the energy flow around the elastic element is one of the keys to improving the energy efficiency of \VSAs.}
In general, the power flow of this element can be represented in the form of power conversion $\powerin = \powerout + \dot{E_s}$, where $\powerin, \powerout$ is the power drained and delivered by the compliant actuation module, respectively, and $\dot{\Es}$ is the rate of change of energy stored. 
As shown in the diagram, in the types of \VIAs considered in this paper,  $\powerin$ consists of power input from the two motors ($\powerinone$ and $\powerintwo$), hence 
\begin{align}
\nonumber\dot{E_s} &= \frac{\partial E_s}{\partial q} \dot{q} + \frac{\partial E_s}{\partial \theta_1} \dot{\theta}_1 + \frac{\partial E_s}{\partial \theta_2} \dot{\theta}_2 \\ \nonumber
&= -\tau_s \dot{q} - \tau_{1}\dot{\theta}_1 - \tau_{2}\dot{\theta}_2 \\ 
&= - \powerout + \underbrace{\powerinone + \powerintwo }_{\powerin}.
\label{e:power_flow}
\end{align}
$\powerout$ is bi-directional which means that the elastic element can deliver energy to, or receive energy from, the link that, in turn, exchanges energy with the environment via interaction.

Regulating the energy flow around the elastic element, as governed by \eref{power_flow}, is one of the keys to improving the energy efficiency of compliant actuators. The majority of prior work in this area has focused on this issue, essentially prioritising the problem of \textit{energy consumption}. In this, two broad categories of approach can be identified:
\il{%
	\item exploiting energy storage $E_s$ and 
	\item reducing energy cost of stiffness modulation $\powerintwo$.
}

\subsubsection{Exploiting energy storage}
Energy storage occurs when $\powerin>\powerout$, and release occurs when $\dot{E_s}<0$ contributing positive output power $\powerout$. One of the appealing features of \VIAs is that they can build up a reserve of energy in the elastic element by receiving power from motors, or through interactions with the environment, and time its release according to task demands.
%

Following this idea, numerous mechanisms and control schemes for physically compliant actuation have been developed 
considering problems of periodic movements (\eg walking \cite{Collins2005icra,Vanderborght2006,VanderBorght2008_bipedal,Hutter2013}, lower-limb prosthetics \cite{Au2007} and cyclic manipulations \cite{Haddadin2011,Lakatos2014,Velasco2015,Matsusaka2016,Haddadin2018}) and discrete movements (\eg throwing \cite{Braun2012,Ozparpucu2013}). 
For example, Vanderborght \etal \cite{Vanderborght2006,VanderBorght2008_bipedal} proposed to exploit natural dynamics in bipedal locomotion by fitting the compliance of the actuator to that of the desired trajectory. 
Matsusaka \etal \cite{Matsusaka2016} used resonance-based control to find the energy-optimal constant stiffness for a cyclic pick-and-place task. 
Haddadin \etal \cite{Haddadin2018} developed a controller to dribble a ball stably using minimal peak power based on analysis of the stability of limit cycles and the effects of hand stiffness for robustness and energy efficiency. 

Optimal control has been shown to exploit the energy storage effect in both strictly periodic \cite{Nakanishi2011,Velasco2015} and discrete tasks. For instance, Braun \etal \cite{Braun2012} showed that discrete movements like throwing can be optimised by gradually feeding energy into the elastic elements before releasing it explosively for the throw, thereby amplifying the instantaneous power output, beyond what would otherwise be possible with the motors. 

%

\subsubsection{Reducing energy cost of stiffness modulation}
The benefits of energy storage can only be enjoyed if there is efficient power flow between the motors and the eleastic element (see \fref{model_and_powerflow}\ref{f:power-flow}), however, in the early development of \SEA s and \VSAs it was observed that the adjustment of stiffness causes high energy consumption. Even when $\powerintwo=0$, the motor may still be consuming energy to maintain the elongation or compression of the elastic element. This has motivated several studies into energy-efficient design of stiffness modulation mechanisms in \VIAs. For instance, Jafari \etal \cite{Jafari2015} suggested a lever mechanism for adjusting stiffness and Braun \etal \cite{Braun2016} proposed a minimalistic stiffness modulator, both of which avoid having motor drives work against spring forces. Parallel springs are implemented in \cite{plooij2016} and \cite{JimenezFabian2017} to reduce required torque by locking potential energy into the parallel springs.


\subsubsection{Energy efficiency through regeneration}
The above examples addressed the energy efficiency in terms of \emph{energy consumption} via either design or control. An alternative is to focus on \textit{energy regeneration}. From \fref{model_and_powerflow}\ref{f:power-flow}, it can be seen that the link dissipates energy via damping elements, a uni-directional flow. If the latter primarily consist of frictional elements, this energy is wasted. However, if the damping mechanism is such that the dissipated energy can be harvested, this energy has the potential to be used to recharge the power source and decrease the overall net consumption. So far, this possibility has received little attention.

\label{s:related_work-regenerative_braking}
For example, \cite{Seok2015Cheetah} implemented regenerative electric motor drivers on the MIT Cheetah, enabling the motors to be used to both actuate and brake the joint, however, that actuation system is based on active impedance control, not physically compliant \VIAs. The same regenerative braking principle is used as a kinetic energy harvester on a lower limb exoskeleton \cite{Donelan2008}, however, there the motor is used purely as a generator and plays no role as an actuator.
Radulescu \etal \cite{Radulescu2012} showed the role that DC motor damping can play in generating braking force in a \VIA, but did not explore its ability to harvest energy. 

It should be noted that regenerative braking technology has been widely used in vehicles driven by electric motors \cite{Hellmund1917,gao1999investigation} or equipped with regenerative suspension systems \cite{Zhang2018_shock}. However, the requirements for \VIAs\ are different from such use cases. Firstly, for general purpose compliant actuators, the movement is typically complex and bidirectional, whereas in vehicles and the locomotion problems considered above,
the braking force required is basically unidirectional. Secondly,  regenerating energy in \VIAs with variable physical damping couples the joint dynamics with the efficiency. Energy regeneration assigns an additional role for variable damping beyond braking and joint stabilisation, thus more investigation is needed to determine appropriate control strategies, in order to balance the trade-off between optimality of energy cost/regeneration and task achievement for specific tasks. 

The next section describes how both issues can be addressed by \il{\item proposing a damping module design capable of harvesting energy from bidirectional movements, and \item evaluating optimal control as a means for dealing with the energy/task performance trade-off.}

\subsection{Dynamic and regenerative braking}\label{s:background-braking}\noindent
Among the different methods of implementing variable physical damping in \VIAs, \textit{damping by motor braking} presents the greatest promise for incorporating energy harvesting by utilising the regenerative braking technique. For this, two main approaches are available, namely \il{\item \emph{dynamic braking} and \item \emph{regenerative braking}}. In both cases, the back electromotive force is used to resist movement proportional to the effective resistance of the damper motor circuit, causing a variable damping effect. 
\begin{figure*}[t]
	\centering
\begin{overpic}[scale=1]{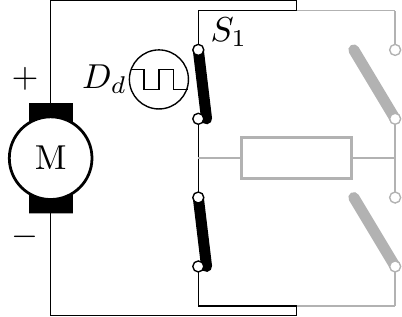}\put(0,5){\ref{f:scheme-dynamic}}\end{overpic}\hfill
\begin{overpic}[scale=1]{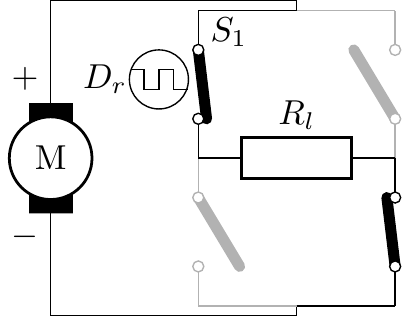}\put(0,5){\ref{f:scheme-rege}}\end{overpic}\hfill
\begin{overpic}[scale=1]{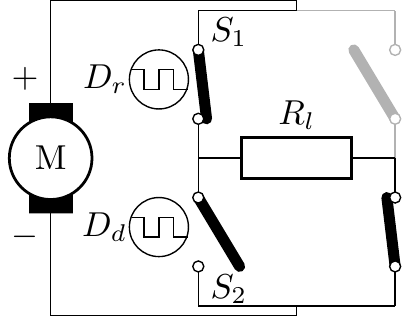}\put(0,5){\ref{f:scheme-hybrid}}\end{overpic} \hfill
\begin{overpic}[scale=1]{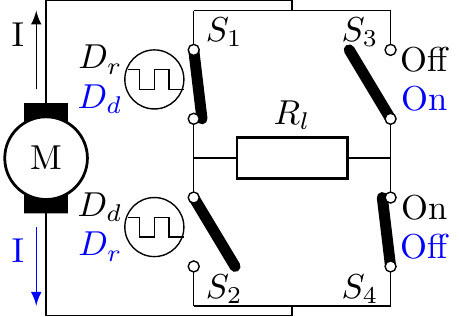}\put(-5,5){\ref{f:scheme-bidirection}}\end{overpic}\hfill
\vspace{0.4cm}
\begin{overpic}[scale=0.78]{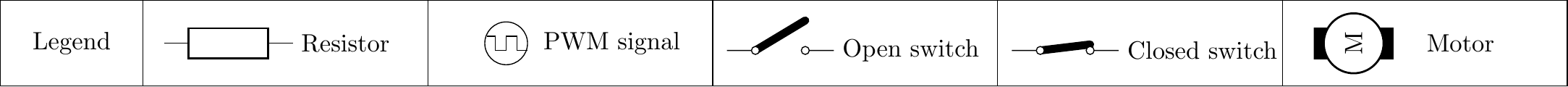}\end{overpic}%

	\caption{\label{f:schemes}Conceptual diagrams of 
\begin{enumerate*}[label=(\alph*)]
\item\label{f:scheme-dynamic} dynamic, 
\item\label{f:scheme-rege} regenerative,
\item\label{f:scheme-hybrid} hybrid dynamic-regenerative, and
\item\label{f:scheme-bidirection} bidirectional hybrid dynamic-regenerative 
\end{enumerate*} 
 braking circuits. In the circuits, $S_i, i=1,...,4$ are switches and $D_r, D_d$ are pulse-width modulated (PWM) control signals. The diagrams are sketched in a synthesised way to show how the different functions can actually be realised in a single circuit (in diagrams \ref{f:scheme-dynamic}-\ref{f:scheme-hybrid}, the parts of the circuit shown in grey should be considered disabled or disconnected).  Generally, the circuit consists of an energy storage element, modelled by a resistor, connected to a DC motor as a damper, with a switching mechanism. In diagram \ref{f:scheme-bidirection}, the current $I$ can have two directions indicated by black and blue arrows. When it flows in the direction indicated by the black arrow, $S_3$ is set to off and $S_4$ is on; the PWM signals for $S_1,S_2$ are control variables $D_r,D_d$ respectively. When the current flow is reversed in the direction of blue arrow, the control variable used for $S_1$ is switched to $D_d$ and $D_r$ controls $S_2$; the on-off modes of $S_3,S_4$ are switched as well.}
\end{figure*}
\subsubsection*{\schemeref{dynamic-braking} - Dynamic braking}\manuallabel{scheme:dynamic-braking}{1}
Dynamic braking in the context of \VIA\ design was first proposed by \cite{Radulescu2012}. A circuit diagram for this scheme is illustrated in \fref{schemes}\ref{f:scheme-dynamic}. 
In this mode, the damping effect is modulated by changing the duty-cycle $\DCd$ that controls the portion of time that a switch $S_1$ spends in the open or closed position, thereby altering the effective resistance of the circuit. The damping coefficient follows the equation
\begin{equation}
	d = \frac{n_d^2 k_t^2 \DCd}{R_m}= \bar{d}_1 D_d \label{e:d-scheme-dynamic}
\end{equation} 
where $n_d$ is the gear ratio of damping motor, $k_t$ is the torque constant and $R_m$ is the resistance of the motor. Note that, since $0\le \DCd \le1$, the maximum damping coefficient that can be provided by dynamic braking is $\bar{d}_1=n_d^2 k_t^2/R_m$. 

In energy terms, dynamic braking is effective 
since it dissipates kinetic energy of the output link as heat in the electrical circuit. It does not, however, charge energy to any electrical source, so the regeneration power is zero ($\Prege = 0$). In other words, this (potentially useful) energy is simply discarded, reducing the overall energy efficiency of the system.

\subsubsection*{\schemeref{regenerative-braking} - Regenerative braking}\manuallabel{scheme:regenerative-braking}{2} 
Regenerative braking refers to the situation where the power generated by the motor through kinetic motion of the output link is used to recharge an electrical storage element (\eg battery, supercapacitor). To implement regenerative braking, the electrical storage element can be simply connected to the circuit of the damping motor, as shown in \cite{Donelan2008}. In the context of \VIA\ design, this can be implemented through the circuit in \fref{schemes}\ref{f:scheme-rege}. 

In regenerative braking mode, the damping effect is dependent on the combined effective resistance of the circuit containing the electrical storage element. Similar to dynamic braking, this can be modulated by controlling the duty-cycle $\DCr$ of a switch. The damping coefficient and the regeneration power can be calculated as
\begin{equation}
	d = \frac{n_d^2 k_t^2 \DCr}{R_m + R_l} = \bar{d}_2 \DCr \label{e:d-scheme-rege}
\end{equation}
and
\begin{equation}
	\Prege = \frac{R_l n^2_d k^2_b \dot{q}^2 D_r}{(R_m + R_l)^2}=\alpha \bar{d}_2 \dot{q}^2 D_r, \label{e:Prege-scheme-rege}
\end{equation}
respectively, where $R_l$ is the internal resistance of the electrical storage element (\eg a battery) and $\alpha = R_l/(R_m+R_l)$. $k_b$ is the back-EMF constant and is equal to $k_t$.

Note that, introducing regenerative braking means that the mechanical energy that is otherwise discarded in the dynamic braking scheme can be harvested, enhancing the overall energy efficiency of the system. However, note also that, compared to dynamic braking, the maximum damping coefficient that can be produced by regenerative braking, $\bar{d}_2=n_d^2 k_t^2/(R_m+R_l)$, is \emph{decreased} since adding an electrical load for charging increases the total equivalent resistance of the circuit. This can be a drawback in applications where higher levels of damping are needed (\eg when there is need for a high dynamic response and therefore heavy braking of rapid movements). Another issue is that, since the electrical storage element is usually unidirectional, the current in \schemeref{regenerative-braking} has to be unidirectional as well. In order to deal with current following in both directions resulted from the bidirectional movements (which is common in robotic applications), a reversing mechanism is needed. One solution for this is to introduce another switch, as illustrated in \fref{schemes}\ref{f:scheme-bidirection}---this will be introduced in detail in \sref{hybrid-damping}.

\section{Hybrid dynamic-regenerative braking}\label{s:hybrid-damping}\noindent%
To meet the requirements for the regenerative damping system to be used with \VIAs, this section presents a variable damping scheme---termed \emph{hybrid braking}---that switches between dynamic braking (\schemeref{dynamic-braking} as described in \sref{background-braking}) and pure regenerative braking (\schemeref{regenerative-braking} in \sref{background-braking}) to achieve the benefits of both. Note that, the proposed scheme is designed as a standalone module to be used in conjunction with a variable stiffness mechanism, which is not restricted to the example taken in this paper (where the joint is connected in series with a variable spring to a motor, and the variable spring is controlled by a second motor to adjust the joint stiffness), but could be agonistic/antagonistic employment of springs, series elastic actuators (SEAs), or other complex arrangement that cannot be represented by the model in \fref{model_and_powerflow}\ref{f:mass-spring-damper}. It could, in theory, also be applied to any joint with high compliance or backdrivability, even when the drive motor is rigidly connected to the joint.

\subsection{Hybrid damping circuit}\noindent%
The hybrid damping scheme proposed here is implemented through the circuit depicted in \fref{schemes}\ref{f:scheme-hybrid}. It uses two switches (denoted $S_i$, $i\!\in\!\{1,2\}$) that switch at high frequency between \il{\item pure regenerative braking, and \item a blend of dynamic and regenerative braking}. The principle by which the proposed scheme operates is as follows. 

When switch $S_2$ is open, the module acts in regenerative braking mode, whereby current flows through the power storage element, with the effective resistance (damping level) determined by the duty cycle of $S_1$. (Note that, this results in an equivalent circuit to that used in \schemeref{regenerative-braking}, \cf \fref{schemes}\ref{f:scheme-rege}.) On the other hand, when $S_1$ and $S_2$ are closed, there is a short circuit that causes current to bypass the resistive load $R_l$, creating a dynamic braking effect. In this case, the damping level can be determined by keeping $S_1$ closed and modulating the duty cycle of $S_2$. This enables a third braking scheme to be defined, alongside Schemes \ref{scheme:dynamic-braking} and \ref{scheme:regenerative-braking},
as follows.

\subsubsection*{\schemeref{hybrid-braking} - Hybrid braking}\manuallabel{scheme:hybrid-braking}{3} 
When the required damping $d^*$ is small enough, \ie $d^*\le \bar{d}_2$, it can be provided by pure regenerative braking, so $S_2$ is opened ($\DCd=0$). When the required damping is greater, \ie $d^*>\bar{d}_2$, $S_1$ is closed ($\DCr=1$) and $\DCd$ is used to control $S_2$ to blend dynamic and regenerative braking. 

The resulting damping coefficient and regeneration power are:
\begin{align}
	d &= \bar{d}_2 \DCr + \alpha \bar{d}_3 \DCd \label{e:d-scheme-hybrid0}\\
	\Prege &= \alpha \bar{d}_2 \dot{q}^2 (\DCr - \DCd)=\Prege^0\dot{q}^2\label{e:Prege-scheme-hybrid}
\end{align}
where $\Prege^0 = \alpha \bar{d}_2 (\DCr - \DCd)$. Note that, if $\DCr=\DCd=1$, the same maximum damping coefficient as that achievable in a pure dynamic braking can be achieved, \ie $\bar{d}_3=\bar{d}_1$. This, however, comes at the cost of the regeneration power vanishing ($\Prege=0$). 

\subsection{Hybrid Damping Control Modes}\label{s:control-modes}\noindent
In principle, each of the switches in the proposed circuit may be independently controlled by its own duty-cycle. While this enhances the flexibility of the damping module, it also introduces an undesirable layer of complexity to its control. 
\begin{figure}[!htb]
	\begin{overpic}[width=\linewidth]{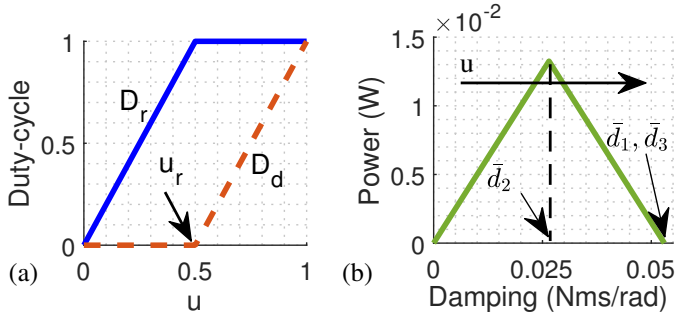}\put(0,5){\ref{f:D1D2-u}}\put(50,5){\ref{f:Prege-damping}}\end{overpic}%
	\caption{\label{f:D1D2-u_Prege-damping} Hybrid damping control modes.
		\begin{enumerate*}[label=(\alph*)]
			\item\label{f:D1D2-u} Mapping from control input $u$ to duty cycles $D_r, D_d$.
			\item\label{f:Prege-damping} Relation between regeneration power ($\Prege$) and damping.
		\end{enumerate*}
	}
\end{figure}

To address this, and enable the simple control of the module through a single control variable $u \in [0,1]$, the duty cycles of the switches can be coupled through the following relation
\begin{align}
	D_r &= \left\{
	\begin{aligned}
	\frac{u}{u_r}, u \leqslant u_r\\
	1, u > u_r
	\end{aligned}
	\right. \nonumber \\
	D_d &= \left\{
	\begin{aligned}
	0, u \leqslant u_r\\
	\frac{u - u_r}{1 - u_r}, u > u_r
	\end{aligned}
	\right.\label{e:D1D2}
\end{align}
where $u_r$ corresponds to the maximum damping coefficient of regenerative braking ($d(u_r) = \bar{d}_2$) and depends on the user's selection. In this paper, $u_r$ is chosen to be $0.5$. Substituting \eref{D1D2} into \eref{d-scheme-hybrid0}, the damping coefficient as a function of $u$ is simplified to
\begin{equation}
	d(u) = \bar{d}_3 u.\label{e:d-scheme-hybrid}
\end{equation}
As illustrated in \fref{D1D2-u_Prege-damping}\ref{f:D1D2-u}, 
when $u\leqslant 0.5$, $D_d$ remains at zero (\ie switch $S_2$ is open) and $D_r$ is linearly mapped from $u \in [0,0.5]$ to $[0,1]$, while when $u > 0.5$, $D_r$ is held at unity ($D_r=1$ so $S_1$ is closed) and $D_d$ is linearly mapped from $u \in [0.5,1]$ to $[0,1]$. 

The relation between the damping coefficient $d$ and the power regeneration $\Prege$ for a fixed angular velocity is shown in \fref{D1D2-u_Prege-damping}\ref{f:Prege-damping}. As can be seen, the relationship is non-monotonic and there is a peak for $\Prege$ when $d = \bar{d}_2$, \ie at the upper boundary of the pure regenerative braking domain.

\subsection{Bidirectional damping}\noindent
The hybrid damping circuit described so far enables the modulation of damping force associated with unidirectional motion of the output link. In order to realise damping of bidirectional motion (as is common in many robotic applications), it is necessary to ensure that the current generated by the damping motor always flows into the positive terminal of the electrical storage element. This can be achieved by a four-switch design of the damping circuit, as illustrated in \fref{schemes}\ref{f:scheme-bidirection}. When the current flows from the positive terminal of the damping motor (as shown by the black arrow in  \fref{schemes}\ref{f:scheme-bidirection}), $S_3$ is open and $S_4$ is switched on. When the current flows from the negative terminal of the motor (as shown by the grey arrow), $S_3$ is closed and $S_4$ is open, and $S_1$ is controlled by $D_d$ and $S_2$ is controlled by $D_r$. 

It should be further noted that, this latter circuit, implements the (bidirectional versions of) the two damping schemes outlined in \sref{background-braking} as special cases. For example, \il{\item holding $S_2,S_3$ open, $S_4$ closed and varying the duty cycle of $S_1$ results in regenerative braking, while \item holding $S_3,S_4$ open, $S_1$ closed, and varying the duty cycle of $S_2$ results in pure dynamic braking}. In other words, the same hardware can be used to realise all three damping schemes.
In the following sections, for brevity, the term regenerative damping will be used to refer to the proposed hybrid damping scheme in the context of \VIAs.

%
%


\subsection{Physical realisation of the damping module}\label{s:experiment}\noindent
This section presents the physical realisation of the hybrid damping circuit design introduced above and an experiment to verify the theoretical predictions about the damping/regeneration performance trade-off.
%
\begin{figure*}[!t]
	\centering
	\begin{overpic}[scale=1]{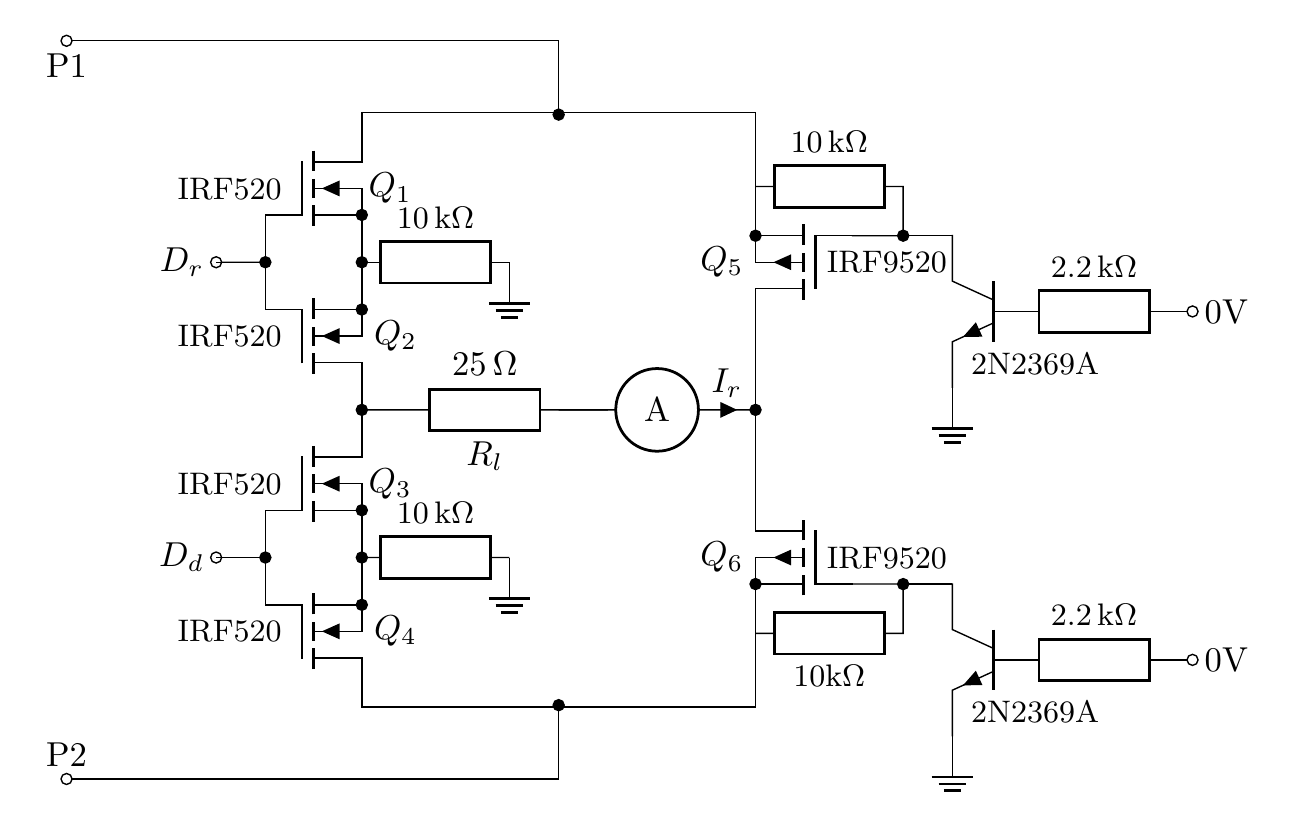}\put(50,2){\ref{f:hw-circuit}}\end{overpic}\hfill
	\begin{overpic}[scale=1]{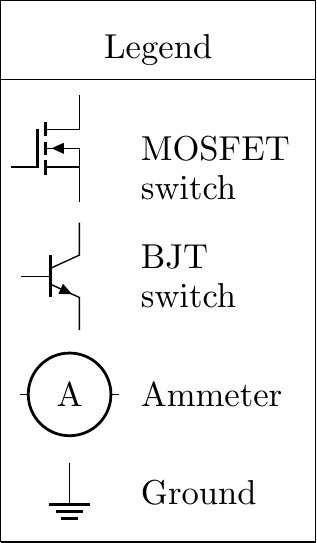}\end{overpic}\hfill
	\vspace{4mm}\hfill
	\begin{overpic}[scale=1]{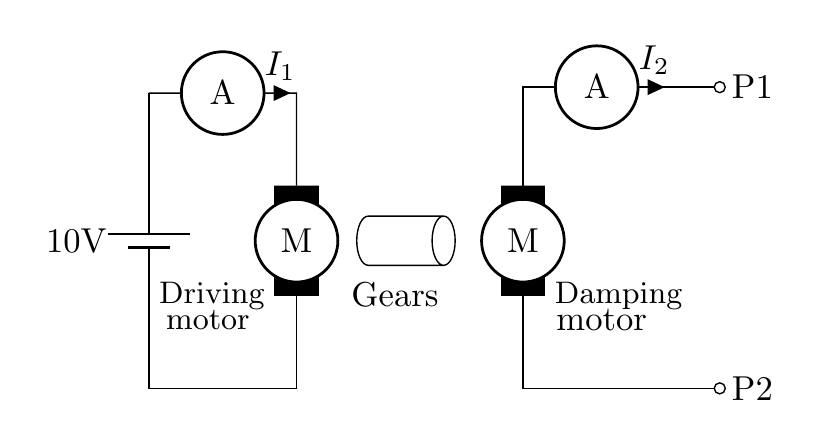}
		\put(50,0){\ref{f:hw-motors-diagram}}
	\end{overpic}
	\begin{overpic}[width=0.45\linewidth]{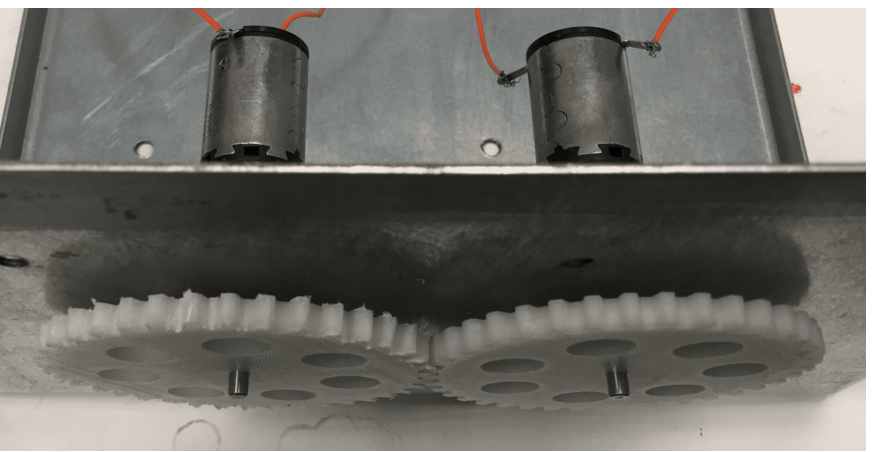}
		\put(20,25){\color{red} Driving Motor}
		\put(60,25){\color{red} Damping Motor}
		\put(50,2){\ref{f:hw-motors}}
	\end{overpic}\hfill
	\caption{\label{f:hardware-setup}
		{Damping and power regeneration measurement experiment setup. Shown are
		\begin{enumerate*}[label=(\alph*)]
			\item\label{f:hw-circuit} circuit diagram of the damping module,
			\item \label{f:hw-motors-diagram} diagram of the experiment setup, and
			\item\label{f:hw-motors} the test rig.
		\end{enumerate*} The terminals of the damping motor are connected to nodes P1 and P2 of the control circuit.}
	}
\end{figure*}
%
The experimental set up is shown in \fref{hardware-setup}. As a simple test-rig, two identical DC motors
(Maxon A-max 22/110125) are coupled through a pair of spur gears to enable one motor (driver) to drive the other (damper), see \fref{hardware-setup}\ref{f:hw-motors-diagram} and \ref{f:hw-motors}. The two motors have the same gearhead with $n_d=20$. The torque constant is $k_t = 0.0212 \mathrm{Nm/A}$ and the motor resistance $R_m=21.2\mathrm{\Omega}$. 

The damper motor is connected to the circuit depicted in \fref{hardware-setup}\ref{f:hw-circuit} (via the nodes P1 and P2), that is the physical realisation of the conceptual diagram \fref{schemes}\ref{f:scheme-bidirection}. In this circuit design, a pair of N-channel MOSFETs (IRF520) is used as one switch to make sure that the switching mechanism works properly for bidirectional current. In \fref{hardware-setup}\ref{f:hw-circuit}, the pair of $Q_1,Q_2$ works as the switch $S_1$, and $Q_3,Q_4$ make the switch $S_2$. Two P-channel MOSFETs (IRF9520) with BJTs (2N2369A) are used as switches $S_3,S_4$. The duty-cycles $D_r,D_d$ are controlled by PWM signals from an Arduino Mega2560 board. By setting $0\mathrm{V}$ signals on the control pins for $Q_5,Q_6$, they are open for just one current direction but closed for the other. For the ease of power measurement, a resistor is used to represent the electrical load ($R_l = 25.3 \mathrm{\Omega}$).

In the experiment, the driving motor is used to drive the system while the damping applied by the second motor is varied, and the resultant motion (motor speeds and energy regeneration) is recorded. Specifically, the driving motor is powered at 10V ($V_{bb} = 10 \mathrm{V}$) constantly by a laboratory DC power supply while the damping motor control input $u$ is varied from 0 to 1 in increments of 0.1 (with the corresponding duty-cycles $\DCr,\DCd$ computed by \eref{D1D2}). Simultaneously, three multimeters (Rapid DMM 318) are used to measure the currents $I_1,I_2,I_r$ through the driving motor, damping motor and the electrical load $R_l$ (represented by a resistor) respectively. The latter data are used to compute the angular speed of the motors $\omega$ and the damping torque $\tau_d$ according to
\begin{align}
V_{bb} &= I_1 R_m + n_d k_t \omega \\
\tau_d &= n_d k_t I_2 = d(u) \omega.
\end{align}
The damping coefficient $d(u)$ for a given $u$ is then estimated by
\begin{equation}
\hat{d}(u) = \frac{n_d^2 k_t^2 I_2}{V_{bb} - I_1 R_m}
\end{equation}
and the regeneration power (normalised by the square of speed for comparison) is estimated as
\begin{equation}
\hat{P}_\mathrm{rege}^0 = \frac{n_d^2 k_t^2 I_r^2 R_l}{ (V_{bb} - I_1 R_m)^2 }.
\end{equation}
The results based on the data collected from $10$ repetitions of the experiment is plotted in \fref{circuit_exp_result} alongside the theoretical predictions (from \sref{control-modes}). 
\begin{figure}[!htb]
	\centering
	\begin{overpic}[width=\linewidth]{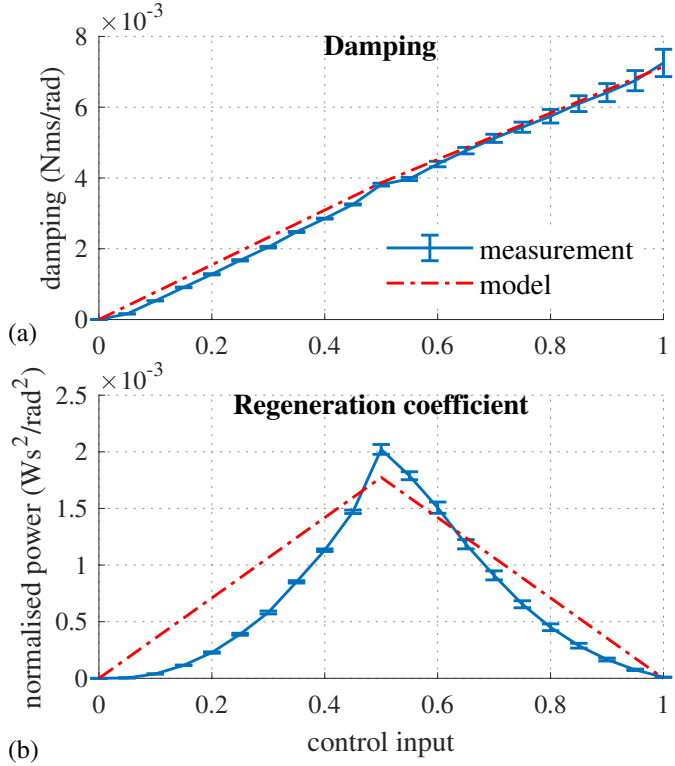}\put(0,55){\ref{f:hw-damp-coeff}}\put(0,0){\ref{f:hw-regen-power}}
	\end{overpic}
	\caption{Results of the damping test experiment. The
	\begin{enumerate*}[label=(\alph*)]
			\item\label{f:hw-damp-coeff} damping coefficient, and
			\item\label{f:hw-regen-power} regeneration coefficient
		\end{enumerate*} for each tested control input $u \in [0,1]$ are shown. The blue error bars represent the means and standard deviations of data points for $10$ repetitions of the experiment and the red line shows the values predicted by the model.}
	\label{f:circuit_exp_result}
\end{figure}

It can be seen that, the experimental data (blue lines in \fref{circuit_exp_result}) is in good agreement with the model predictions (red lines in \fref{circuit_exp_result}). By increasing $u$ from 0 to 1, the damping coefficient $d$ increases almost proportionally. Furthermore, it is verified that, when fixing the angular speed ($\Prege$ has been normalised to estimate $\Prege^0$), the relation between $\Prege^0$ and $u$ is non-monotonic with a peak found at $u = 0.5$. However, the experimental data indicates that for both regions ($u \in [0,0.5]$ and $u \in [0.5,1]$), the regeneration coefficient is not linearly dependent on the control input. This modelling error might be due to unmodelled effects such as circuit inductance and switching frequency.

\section{Simulation and evaluation with ideal VIA}\label{s:toy-example}\noindent Having verified the feasibility of physically implementing the proposed damping module, it is necessary to evaluate its use in the context of robot control. As noted in \sref{related_work-regenerative_braking}, the dual role of the damping module, both for braking and energy harvesting introduces a trade-off between task performance and energy efficiency. To resolve this, it is proposed to employ optimal control to determine the best damping modulation strategy according to task demands. This section presents an evaluation of such a scheme to control the hybrid braking module in the context of a simple example task of target reaching.

For this, a model of a simple pendulum, subject to viscous friction and actuated by an ideal \VIA\ is used
\begin{align}
m l^2 \ddot{q} + b \dot{q} = k(u_2)( u_1 - q) - d(u_3).
\label{e:simple-pendulum}
\end{align}
Here, for simplicity, $m=1 \mathrm{kg}, l = 1 \mathrm{m}$, $b=0.01 \mathrm{Nms/rad}$. The motor positions $\theta_1,\theta_2$ are assumed to be directly controlled by control inputs $u_1,u_2$. $u_3$ is the control input for damping $d$. The right hand side of \eref{simple-pendulum} is the joint torque applied by the ideal \VIA, $u_1 \in [-\pi/2,\pi/2]\,\mathrm{rad}$ controls the equilibrium position and the stiffness $k(u_2)$ is proportional to the control input $u_2\in[0,1]$, \ie
\begin{align}
k(u_2) = \bar{k}u_2,
\end{align}
where $\bar{k}=200\,\mathrm{Nm/rad}$ is the maximum stiffness. The damping $d(u_3)$ as a function of $u_3$ is given by \eref{d-scheme-hybrid}. The corresponding power of regeneration $\Prege$ is assumed to be computed by the model introduced in \sref{hybrid-damping}. The parameters\footnote{These parameters are arbitrarily chosen to give response within a second. Experimentation shows the result is not sensitive to these values.} that characterise the variable damping module are selected to be $\bar{d}_3 = 50\,\mathrm{Nms/rad}$, $\bar{d}_2 = 25 \mathrm{Nms/rad}$ and $\alpha = 0.5$.
The control frequency is set to $50\mathrm{Hz}$.

The task is to reach a target $q^* = \pi/3\,\mathrm{rad}$ from the initial position $q = 0\,\mathrm{rad}$ within a finite time $t_f$ as quickly and accurately as possible, while minimising the energy consumption and control effort. This can be described through minimisation of the cost function 
\begin{multline}
J = \int_{0}^{t_f} [w_1 (q(t) - q^*)^2 + w_2 ( u_1(t)-q^* )^2 \\
+ w_3 u_2^2(t) - w_4 \Prege] \dt \label{e:cost-function},
\end{multline}
where $w_1=1000$, $w_2=w_3=1$, $w_4=0.01$ are weighting parameters. These parameters are selected to take account of the different scales of the terms and allow reaching within a second. 
In the cost function, the first term represents the reaching accuracy and drives the plant to reach the target quickly; the second term is used to penalise deviation of the equilibrium position from the target to increase stability; the third term encourages using lower pretension (corresponding to lower stiffness); and the fourth term is used to encourage using the regeneration.

To simplify the analysis, in the below, the command for equilibrium position is fixed at $u_1 = \pi/3$, while the commands for stiffness and damping are allowed to vary. The optimal open-loop control sequence  for the latter is computed through the \newabbreviation{\ILQR}{ILQR} method \cite{Li2004} with the proposed hybrid braking scheme, and the resultant trajectory of the system is computed by simulating the execution of the open-loop command using the 4th Order Runge-Kutta method. 

To evaluate the energy efficiency of the proposed approach, the total mechanical work\footnote{As motors are not explicitly involved in the model, the mechanical work computed here corresponds with the integration of the power delivered onto the plant ($\powerout$ in \fref{model_and_powerflow}\ref{f:power-flow}), not the power from motors side ($\powerin$).} and the total regenerated energy are computed from the resultant trajectories, \ie
\begin{align}
E &= \int_0^{t_f} k(u_1 - q) \dot{q} \dt\\
\Erege &= \int_{0}^{t_f} \Prege(t)\dt,
\label{e:toy-E}
\end{align}
respectively. The net energy cost can be defined as
\begin{equation}
\Enet = E - \Erege.
\end{equation}
The percentage ratio of energy regeneration
\footnote{Note that, for simplicity, it is assumed here that there is $100\%$ kinetic to electric energy transmission efficiency of the DC motor. In practice, losses are likely to occur due to friction and losses in the conversion from the mechanical to the electrical domain.} 
can be computed by
\begin{equation}
\eta = \frac{\Erege}{E}.
\end{equation}

For comparison, the experiment is repeated
with \il{\item pure dynamic braking (\schemeref{dynamic-braking}), \item pure regenerative braking (\schemeref{regenerative-braking}), \item the case where the damping is fixed at the maximum power of regeneration ($d=\bar{d}_2$), and \item\label{i:toy-case-critical} a critically damped system}. In the latter, the stiffness is chosen to be $k=100 \mathrm{Nm/rad}$ and the damping is fixed to $d=20 \mathrm{Nms/rad}$ such that the damping ratio $\zeta=d/2\sqrt{km}=1$. 

\begin{figure}[!htb]
	\centering%
\begin{overpic}[width=\linewidth]{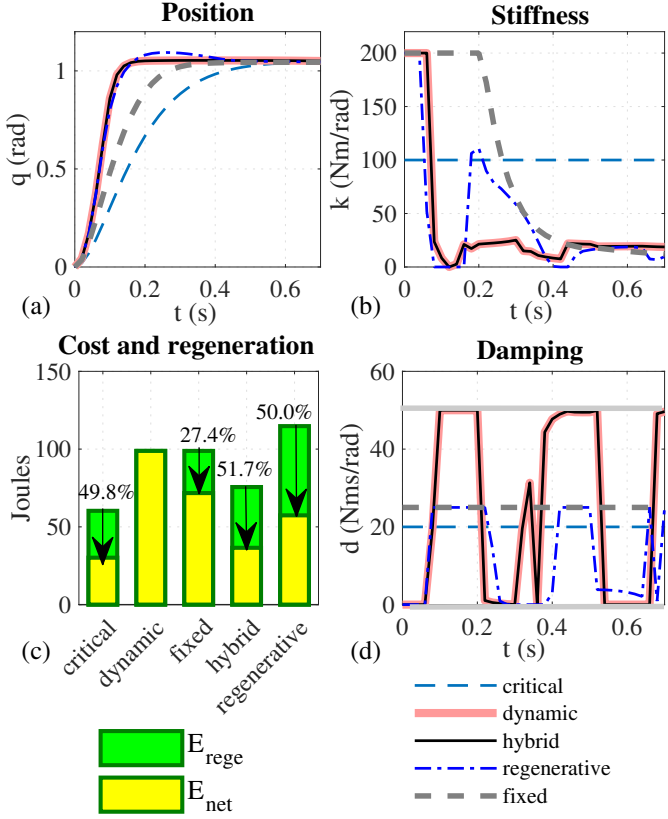}
\put( 2,62){\ref{f:toy-q}}
\put(42,62){\ref{f:toy-k}}
\put( 2, 20){\ref{f:toy-E}}
\put(42, 20){\ref{f:toy-d}}
\end{overpic}%
	\caption{\label{f:toy_example} {Test of reaching task on a simple pendulum with ideal VIA. Shown are optimal
\begin{enumerate*}[label=(\alph*)]
\item\label{f:toy-q} joint angular trajectories,
\item\label{f:toy-k} stiffness profiles,
\item\label{f:toy-E} total mechanical work and percentage ratio of energy regeneration, and
\item\label{f:toy-d} damping profiles for different damping schemes.
\end{enumerate*} The proposed hybrid damping scheme is compared with \il{\item critical damping, \item pure dynamic braking, \item pure regenerative braking, and \item fixed damping ($d=\bar{d}_2$)}.       }
    }
\end{figure}

The results are illustrated in \fref{toy_example}. As can be seen, the trajectory of the critically damped system reaches the target slowly but without overshoot (\fref{toy_example}\ref{f:toy-q}). The system with fixed damping reaches the target quicker than the critically damped one, because it can exploit the variable stiffness. The system with regenerative braking reaches the target quicker still, however, since the damping range is limited in this case, it suffers from overshoot once it reaches the target. In contrast, the dynamic braking and hybrid braking systems reach the target quickest without overshot, so perform best in terms of accuracy. 

Looking at \fref{toy_example}\ref{f:toy-E}, however, it can be observed that the dynamic braking performs worst in terms of net energy cost, since no energy is recovered throughout the movement. This contrasts with the hybrid approach, that achieves fast and accurate movement while also achieving $27.4\%$ energy recovery, thereby lowering the net energy cost.

Overall, the proposed hybrid scheme offers a good trade-off between task accuracy and energy efficiency.

\section{Experiment: Long-term robotic deployment}\label{s:evaluation}\noindent%
In the real-world deployment of compliant robotic systems, \VIAs must be able to withstand many use cycles with unpredictable task demands. This section reports an experiment designed to evaluate the likely effectiveness of the proposed damping scheme in the context of long-term use. For simplicity, the test case chosen is the task of performing consecutive point-to-point reaching movements to a series of random targets generated on the fly (\ie as the robot is moving). The aim is to examine the performance of the proposed scheme as compared to competing ones \emph{where stiffness and/or damping are fixed} against multiple performance metrics. \edit{The latter take into account of task performance (settling time and overshooting of movements), energy consumption and regeneration.}{}
Note that, while such tasks are common in many robotic applications (\eg a robot deployed to tidy a room may have to reach and grasp many objects at uncertain locations), they are challenging from the point of view of energy management, since the movements are non-periodic and unpredictable\edit{, hence, physical factors, such as the natural frequency of the device cannot easily be exploited}{}. The following reports the experimental design and procedure in detail.


\subsection{Hardware specifications}
\begin{figure}[!htb]
	\centering
	
	\begin{overpic}[width=\linewidth]{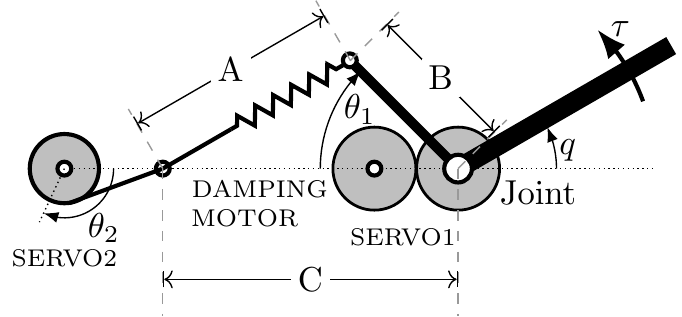}
	\put(0,40){\ref{f:maccepavd-diagram}}
	\end{overpic}
	\begin{overpic}[width=\linewidth]{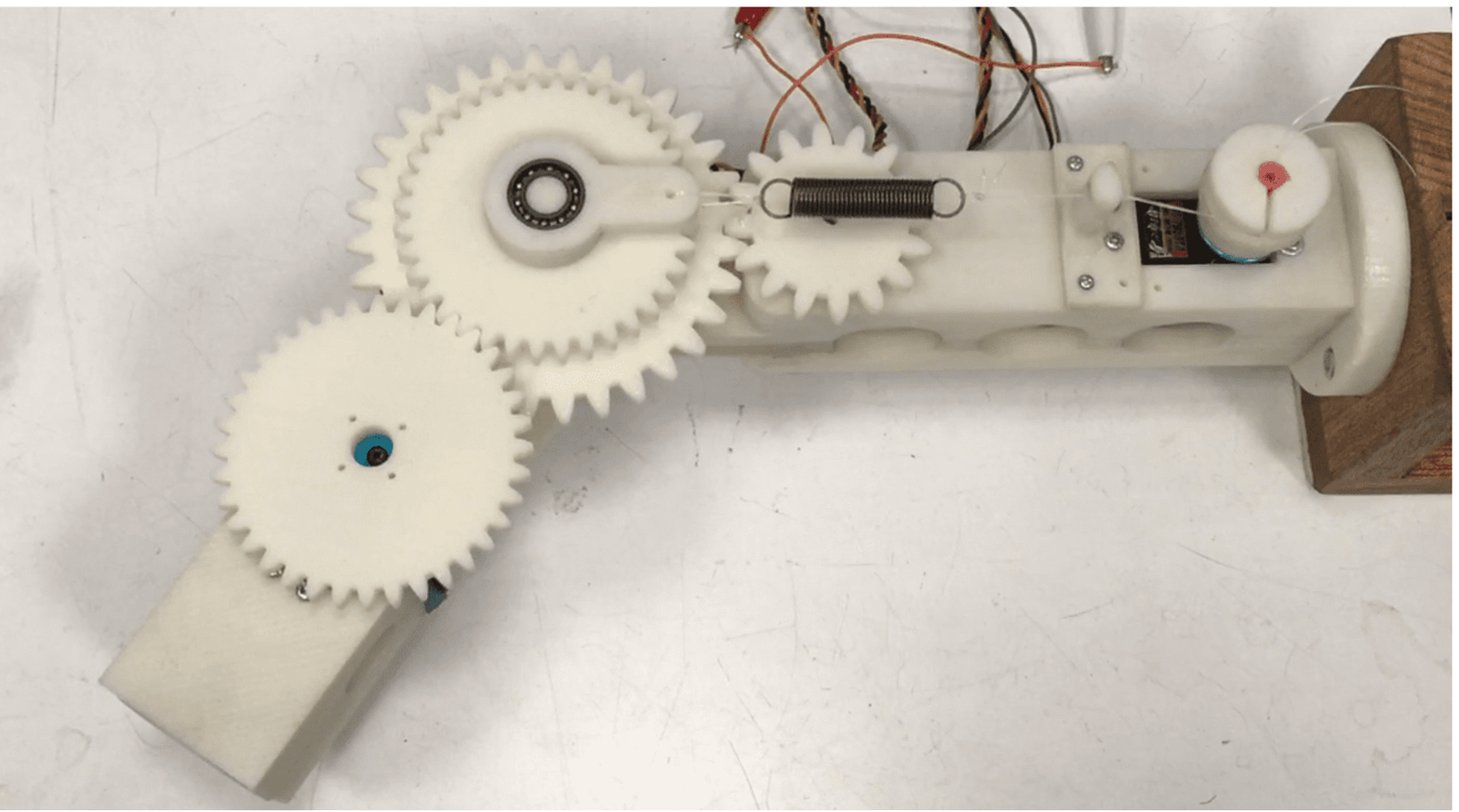}	
	\put(0,50){\ref{f:maccepavd-hardware}}
	\end{overpic}
	\caption{\label{f:maccepavd}\begin{enumerate*}[label=(\alph*)]
	    \item\label{f:maccepavd-diagram} Schematic diagram and
	    \item\label{f:maccepavd-hardware} hardware
	\end{enumerate*} of \protect\MACCEPA\ \cite{VanHam2007} with variable damping \cite{Radulescu2012}. In the results reported here, $B = 3.6\,\mathrm{cm}, C=13.5 \, \mathrm{cm},r = 1.5 \, \mathrm{cm}$ and the spring has linear spring constant $\kappa=394 \,\mathrm{N/m}$. The link has inertia $m=0.0036 \, \mathrm{kg m^2}$ and friction coefficient $b=0.0077 \, \mathrm{Nms/rad}$. The gear ratio between the joint and damping motor is $n_d=40$.}
\end{figure}

\noindent
To evaluate its use, the damping module developed in \sref{experiment} is implemented on a physical \VIA. Specifically, the experimental platform consists of a 3D-printed, single-joint robot using the \newabbreviation{\MACCEPAVD}{MACCEPA-VD} \cite{Radulescu2012,VanHam2007} mechanism for actuation, where the joint stiffness is adjusted by changing spring pretension (see \fref{maccepavd}\ref{f:maccepavd-hardware}). 
A schematic diagram of the system is shown in \fref{maccepavd}\ref{f:maccepavd-diagram} alongside the design parameters. In this implementation of the \MACCEPAVD, the equilibrium position and joint stiffness are controlled by two servomotors (Hitec HS-7950TH and Hitec HSR-5990TG, respectively). A DC motor (Maxon A-max 22/110125) is attached to the joint to serve as the damping motor, whose damping effect is controlled by the control unit introduced in \sref{experiment}. A current sensing module (Adafruit ina219 breakout) is connected in series with the electric load in the circuit to measure the high-side current and voltage to calculate the power of regeneration in real-time. A potentiometer (ALPS RDC503) is used to measure the joint angle. The velocity is then estimated by finite differences on the position data. The software architecture is based on the open-source Robot Operating System (ROS), where the control command is published to a ROS message, which is then subscribed by a microcontroller (Arduino mega2560) to control the servomotors and the damping unit.

\subsection{Control of the variable impedance robot}\noindent
The variable impedance mechanism has intrinsic redundancy in its internal actuation. Optimal control has been demonstrated to be a straightforward and simple way to resolve this redundancy \cite{Braun2013} and efficient numerical solutions are available through local, iterative algorithms, such as \ILQR. In the experiments reported here, \ILQR\ is used to design the control sequence for the robot on the fly, as each reaching target is generated.

\subsubsection{Robot dynamics}
To determine the optimal control sequence, \ILQR\ requires a model of the dynamics of the system. For the \MACCEPAVD, the dynamics are governed by the equations
\begin{align}
    \ddot{q}    &= (\tau_s - d(u_3) \qdot - b\qdot -\tau_{\mathrm{ext}}) \inertia^{-1} \\\label{e:motor-dynamics1}
	\ddot{\theta}_1 &= \beta^2(u_1 - \theta_1) - 2\beta\dot{\theta}_1    \\\label{e:motor-dynamics2}
	\ddot{\theta}_2 &= \beta^2(u_2 - \theta_2) - 2\beta\dot{\theta}_2 
\end{align}
where $q, \dot{q}, \ddot{q}$ are the joint angle, velocity and acceleration, respectively, $b$ is the viscous friction coefficient for the joint, $\inertia$ is the link inertia, $\tau_s$ is the torque generated by the spring force, and $\tau_{\mathrm{ext}}$ is the joint torque due to external loading (the following reports results for the case of no external loading, \ie $\tau_{\mathrm{ext}}=0$). $\theta_1, \theta_2, \dot{\theta}_1, \dot{\theta}_2, \ddot{\theta}_1, \ddot{\theta}_2$ are the motor angles, velocities and accelerations. The motor angles $\theta_1,\theta_2$ are controlled by $u_1 \in [-\pi/3,\pi/3], u_2 \in [0,\pi/3]$ respectively. The servomotor dynamics \eref{motor-dynamics1}, \eref{motor-dynamics2} are assumed to behave as a critically damped system,\footnote{This 2nd order dynamical model is widely used in the literature, \eg in \cite{Braun2013}. Here, the coefficient $\beta$ is chosen empirically to fit the step response of the servomotors.} with $\beta=25$.

The torque $\tau_s$ can be calculated as follows:
\begin{equation}
	\tau_s = \kappa B C \sin{ (\theta_1 - q) } (1+ \frac{r \theta_2 - |C-B|}{A(q,\theta_1)}) 
\end{equation}
where $A(q,\theta_1) =\sqrt{B^2+C^2-2BC\cos{ (\theta_1 - q) }}$, $B$ and $C$ are the lengths shown in \fref{maccepavd}, $r$ is the radius of the winding drum used to adjust the spring pre-tension, and $\kappa$ is the linear spring constant. 
The damping coefficient $d(u_3)$ depends on control input $u_3$ and is
calculated according to the damping scheme used (\ie \eref{d-scheme-dynamic}, \eref{d-scheme-rege} or \eref{d-scheme-hybrid}). Note also that,
the stiffness of this system depends on the joint and motor positions $q,\theta_1,\theta_2$
\begin{multline}
	k(q,\theta_1,\theta_2) =\kappa BC \cos (\theta_1-q) (1+ \frac{r \theta_2
 - |C-B|}{A}) \\
- \frac{\kappa B^2 C^2 \sin^2 (\theta_1 - q (r \theta_2 - |C-B|))}{A^{\frac{3}{2}}}. \label{e:maccepa-stiffness}
\end{multline}
\begin{figure*}[t!]
    \centering
    \includegraphics[width=\linewidth]{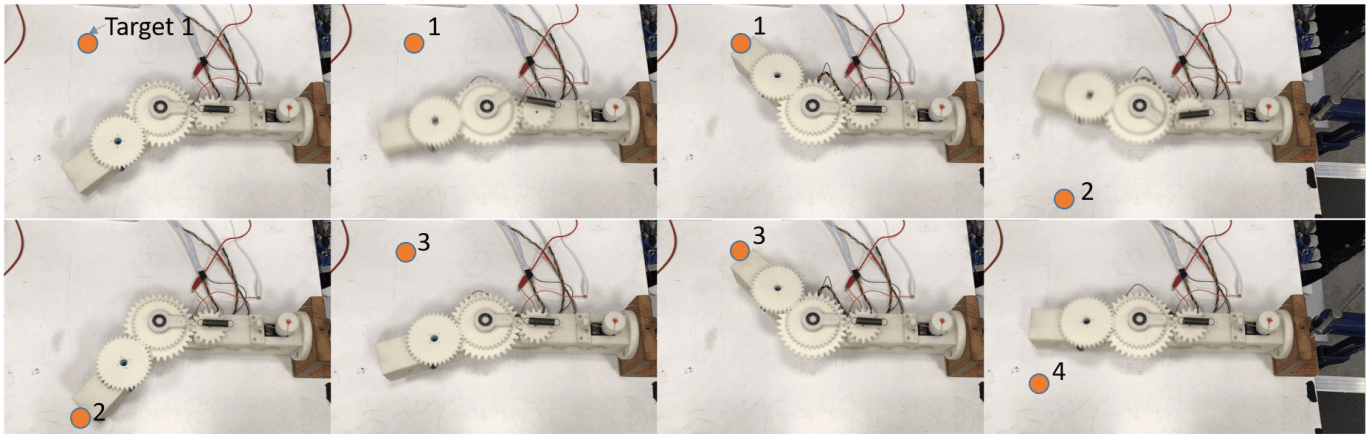}
    \caption{Snapshots of consecutive point-to-point reaching in the VSVD condition. The orange points overlaid show the reaching targets.}
    \label{f:snapshot_consecutive_reaching}
\end{figure*}
So that the same spring pretension $\theta_2$ can result in different joint stiffness under different joint configurations.

\subsubsection{Optimal control formulation}\noindent
To represent the problem as an optimal control problem the dynamics are formulated as a state-space model $\dot{\mathbf{x}} = \mathbf{f}(\mathbf{x},\mathbf{u})$, where $\mathbf{x}=(x_1,x_2,x_3,x_4,x_5,x_6)^\top=(q,\dot{q},\theta_1,\theta_2,\dot{\theta}_1, \dot{\theta}_2)^\top \in \mathbb{R}^6$ denotes the state space vector, $\mathbf{u}=(u_1,u_2,u_3)^\top \in \mathbb{R}^3$ is the control input, and $\mathbf{f}$ is defined as
\begin{align}
    \mathbf{f} = \left\{
    \begin{aligned}
    & x_2 \\
    & (\tau_s(x_1,x_2,x_3) - (d(u_3) + b) x_2 ) \inertia^{-1} \\
    & x_5 \\
    & x_6 \\
    & \beta^2( u_1 - x_3) - 2\beta x_5 \\
    & \beta^2( u_2 - x_4) - 2\beta x_6
    \end{aligned} 
    \right.\label{e:SSM_maccepa}
\end{align}

The task of point-to-point reaching is captured by the cost function
\begin{equation}
    J = \int_0^{t_f}( w_1 ( q - q^* )^2 + w_2 F_s^2 + w_3(u_3-0.5)^2 +w_4\mathbf{u}^\top\mathbf{u} )  \dt
    \label{e:j-consecutive-reaching}
\end{equation}
where $q^*$ is the reaching target, $F_s$ is the spring force, and $t_f$ is the reaching duration (for simplicity, in the experiments reported below, this is fixed at $t_f=1.5$s for each movement). Here, the first term represents the reaching error and has weight $w_1 = 1000$ and the second term penalises the squared spring force ($w_2=1$) which accounts for minimising energy consumption.\footnote{A full model of the energy consumption of this actuator is not available. However, due to its mechanical design (with the spring pre-tension motor working against the spring), the overall electrical power consumption is monotonically related to squared spring force.} The third term penalises deviation of damping control from $0.5$ (since this is known to be the point at which the regeneration coefficient is maximised, see \sref{hybrid-damping}) to encourage energy regeneration (in the experiments reported here it is weighted at $w_3=500$). The last term is added for regularisation of the optimal control solution ($w_4=10^{-6}$). Note that, it is possible to use predicted regeneration power to replace the third term, however, it may result in behaviours sensitive to modelling errors of both dynamics and power regeneration.

\subsection{Consecutive point-to-point reaching experiment}\noindent
The task chosen to evaluate the proposed scheme is consecutive point-to-point reaching to random targets. The experimental procedure is as follows.

A list of $N=25$ locations are generated sequentially as targets for reaching. Each target is drawn uniform randomly (\ie $q^* \sim U[-\pi/3,\pi/3]$), with the minimal distance between the target and the preceding one restricted to be at least $\pi/3$, to exclude very short-range movements. After generation, each target is fed to the cost function \eref{j-consecutive-reaching}, and \ILQR\ is used to determine the optimal control sequence for the movement under the dynamics \eref{SSM_maccepa}, utilising the whole control input space (\ie where all three control variables $u_1$, $u_2$ and $u_3$ are exploited to seek the optimum). The solutions are then executed on the plant, and the resultant joint trajectories and regenerated current are recorded. This procedure is repeated $M=20$ times to get a total of $500$ recorded trajectories for performance evaluation. (In the below, this is termed the \emph{variable stiffness variable damping} (VSVD) condition.) \fref{snapshot_consecutive_reaching} shows snapshots of reaching movements made to a typical sequence of targets.

For comparison, using the same reaching targets, the above procedure is repeated under three further conditions, namely \il{\item \emph{fixed stiffness and fixed damping} (FSFD): a baseline set where $u_1$, $u_2$ and $u_3$ are held at to constant values (in this case, reaching occurs by setting $u_1=q^*$ prior to the onset of each reaching movement), \item \emph{fixed stiffness and variable damping} (FSVD): only the damping control $u_3$ is optimised, while $u_1$ and $u_2$ are held at fixed values, \item \emph{variable stiffness and fixed damping} (VSFD): the damping command $u_3$ is fixed and the equilibrium position and stiffness control inputs are optimised}. In the conditions where the stiffness is fixed (\ie FSFD and FSVD), $u_2$ is set to the minimal stiffness motor angle\footnote{The minimal stiffness motor angle is selected empirically to add sufficient pretension of the spring to provide good reaching accuracy around the zero joint position.} $\pi/6$. In the conditions where the damping is fixed (\ie FSFD and VSFD), the damping command is set at $u_3=0.5$, corresponding to the point at which the regeneration coefficient reaches its maximum.

\subsection{Performance metrics}\noindent
In order to quantitatively compare the results, four metrics are employed to take into account of movement performance as well as energy consumption and regeneration:
\begin{description}
    \item[Settling time] The time when the plant settles down. For a given trajectory, this is defined as the smallest time $t$ where both velocity and acceleration are within the vicinity of zero, \ie $|\dot{q}_t| < \epsilon_1$ and $|\ddot{q}_t| < \epsilon_2$. In our experiments, $\epsilon_1$ was chosen to be approximately $1\%$ of the maximum measured velocity and $\epsilon_2$ was chosen to be $1.5\%$ of maximum measured acceleration.
    \item[Overshoot] The deviation of the joint position from the target point after over-shooting the target. It is defined as the integration of $(q_t - q^*)^2$ from the time at which the target is first reached until the plant settles.
    \item[Energy consumption $\Ein$] computed by integrating $\powerin$ (defined in \sref{power-flow}).
    \item[Regenerated energy $\Erege$] computed by integrating the measured regeneration power.
\end{description}
Each of these are computed using the experimentally recorded data from the robot.
For each trial, the settling time and overshoot of $N$ trajectories are averaged and the energy regeneration and consumption are accumulated.

\subsection{Results}\noindent
\begin{table*}[!ht]
	\centering
	\renewcommand{\arraystretch}{1.1}
	\begin{tabular}{c|c|c|c|c}
		\hline
		Experiment & Settling time (s) & Overshoot ($10^{-2} \mathrm{rad}^{2} \mathrm{s}$) & $\Ein$ (J) & $\Erege$ (J)\\
		\hline
		FSFD & $1.064\pm0.039$ & $4.980\pm1.040$ & $4.080\pm0.552$ & $0.152\pm0.019$ \\
		FSVD & $0.923\pm0.063$ & $1.050\pm0.350$ & $3.989\pm0.541$ & $0.071\pm0.014$ \\
		VSFD & $0.792\pm0.061$ & $0.650\pm1.990$ & $3.412\pm 0.358$ & $0.089\pm0.006$ \\
		VSVD & $0.780\pm 0.045$ & $0.270\pm0.120$ & $3.067\pm0.338$ & $0.092\pm0.010$ \\
		\hline
	\end{tabular}
	\caption{Performance metrics for the four experimental conditions computed on the recorded reaching data. Shown are mean $\pm$ standard deviation of each metric over $20$ trials.}
	\label{t:table_multiple_reach}
\end{table*}
The results for the four experimental conditions are reported in \tref{table_multiple_reach} and their normalised scores are visualised in the radar chart in \fref{longterm_radar}, where the higher the score, the further out the line appears along that dimension (for example, the energy regeneration score, denoted by $\gamma_r$, is the value of $\Erege$ normalised {according to its} the maximum and minimum values in the four experimental conditions, and  $\gamma_t,\gamma_o,\gamma_c$---the normalised settling time, overshoot and energy consumption, respectively---are computed similarly).

\begin{figure}[!htb]
    \centering
    \includegraphics[width=\linewidth]{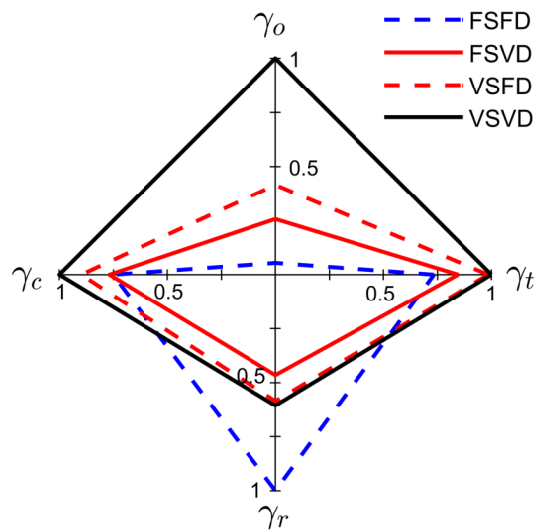}
    \caption{Radar chart showing the normalised reaching scores under the four experimental conditions. The higher the score the better the performance. Shown are mean scores over $20$ trials.}
    \label{f:longterm_radar}
\end{figure}
\begin{figure*}[!htb]
    \centering
    \begin{overpic}[width=\linewidth]{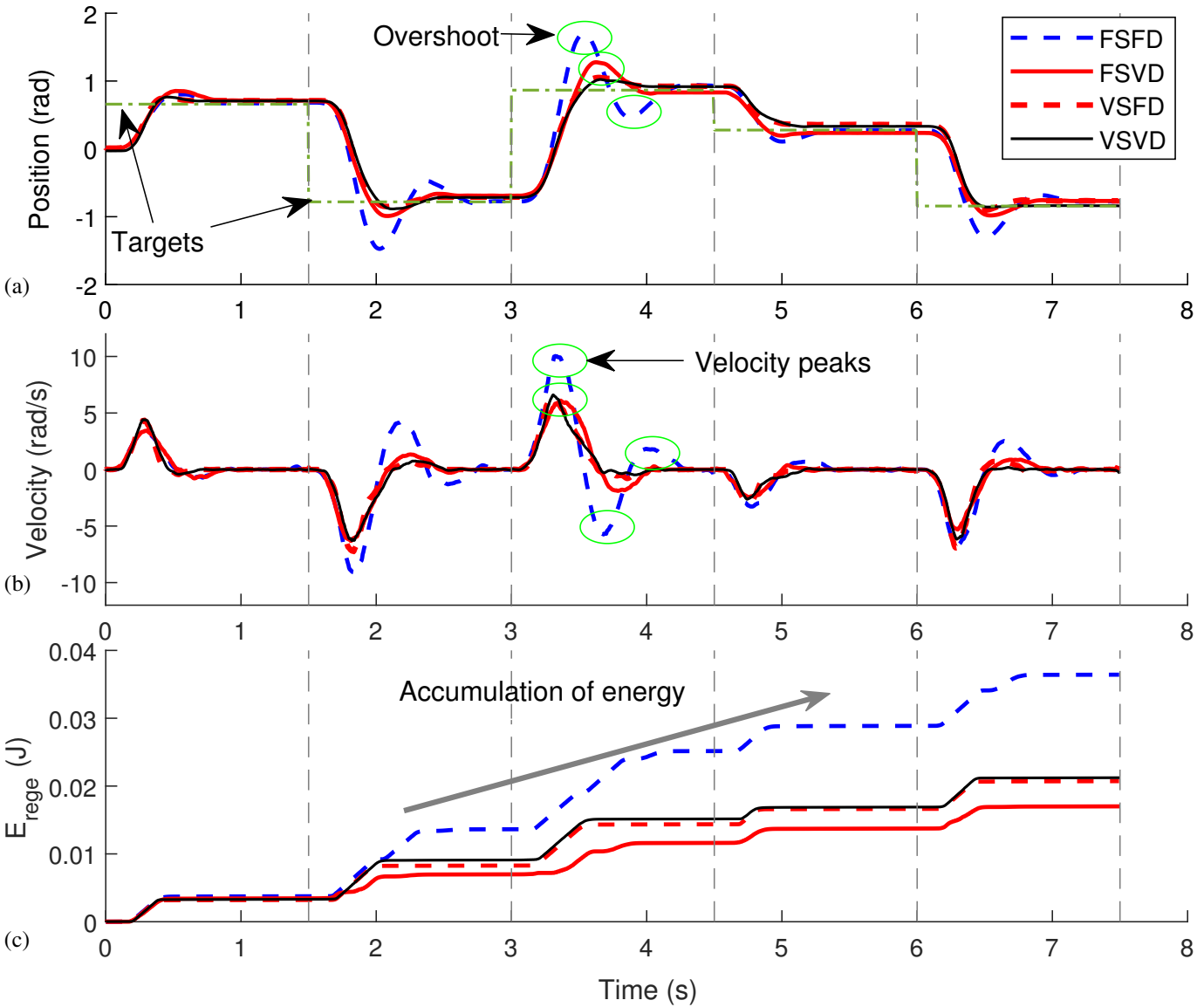}
\put( 0,60){\ref{f:multireach_position}}
\put(0,35){\ref{f:multireach_velocity}}
\put( 0, 5){\ref{f:multireach_energy}}
\end{overpic}%
    \caption{\label{f:multireach} Recorded trajectories 
        \begin{enumerate*}[label=(\alph*)]
\item\label{f:multireach_position} joint position,
\item\label{f:multireach_velocity} velocity, and
\item\label{f:multireach_energy} accumulation of regenerated energy $\Erege$
\end{enumerate*} of five typical examples of consecutive point-to-point reaching movements. The end of each movement is represented by the vertical dashed line. The dot-dashed line in \ref{f:multireach_position} shows the target position for each phase of movement.}
\end{figure*}

Looking at each of these, it can be seen that in all conditions the damping module successfully regenerates power during the movement (see, for example, \fref{multireach}\ref{f:multireach_energy} where a monotonic increase in accumulated energy is seen). Note, however, there is a discrepancy in the amount of energy regenerated and the corresponding performance in the reaching task.

As seen in \fref{longterm_radar}, the baseline condition (FSFD) harvests the most energy, by fixing the damping to the value that provides maximum regeneration coefficient. Although this simple control strategy results in the most energy regeneration in the experiments, it sacrifices movement performance, scoring lowest in terms of the overshoot and settling time. Looking at the trajectory in \fref{multireach} (dashed blue line), it can be seen that there is significant overshoot for multiple targets. The enhanced energy regeneration also does not translate to lower energy consumption (see \fref{longterm_radar} and \tref{table_multiple_reach}). All this suggests that, although the stiffness and damping can be pre-tuned to give good performance for a specific movement, it can only be a solution for a specific task, and thus not suitable for a versatile actuator.

With FSVD, the movement performance in terms of overshoot and settling time is improved compared to FSFD, although the decrease in energy consumption is insignificant. The result confirms that variable damping can be utilised to improve the dynamic performance when the joint has fixed stiffness profile. It gives good overall dynamic performance for varied reaching targets. However, without exploiting variable stiffness, the variable damping cannot ensure energy efficient movements alone. In \fref{longterm_radar} it can be seen that FSVD regenerates the least energy and there is no obvious improvement in terms of energy consumption. 

By modulating the equilibrium position and stiffness VSFD performs moderately better on all performance metrics compared to FSVD. However, when all impedance variables are available to the controller, as in condition VSVD, it can be seen that the performance is significantly improved across the different metrics compared to the other conditions (see \fref{longterm_radar}). The energy efficiency is improved because there is less consumption ($\Ein$) and more regeneration ($\Erege$). Additionally, although the average settling time is almost the same as that of VSFD, there is significantly lower overshoot. 

Overall, these comparisons show that using variable damping in combination with an optimally exploited variable stiffness mechanism can contribute both enhanced dynamic performance and improved energy efficiency (in terms of both consumption and regeneration).


\subsection{Loss of regeneration through over-exploitation of damping}\noindent
Comparing the performance of FSFD and FSVD it can be seen that a relatively modest improvement in reaching performance comes at the cost of a large reduction in the energy regeneration level. This seems surprising since FSVD also has available the possible strategy of keeping the damping fixed (although the damping can be varied, there is no imperative to do so). To better understand this behaviour, it is illuminating to examine in detail the control strategies chosen under the different conditions.

\begin{figure}[!htb]
    \centering
    \begin{overpic}[width=\linewidth]{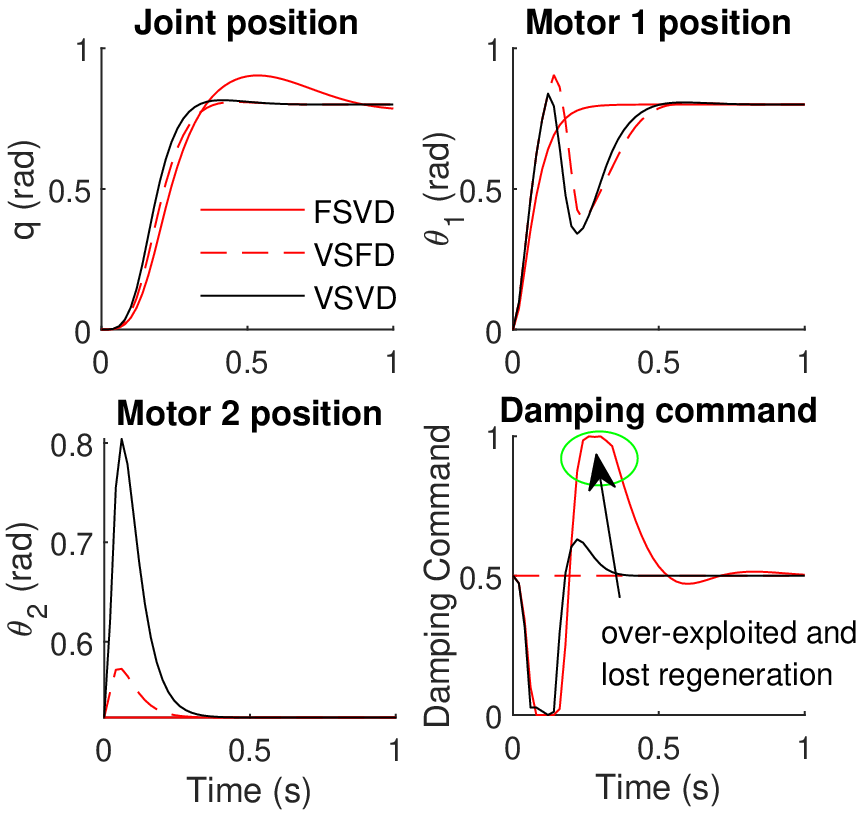}
    \put(0,55){\ref{f:compare_strategies_position}}
    \put(50,55){\ref{f:compare_strategies_motor1}}
    \put(0,5){\ref{f:compare_strategies_motor2}}
    \put(50,5){\ref{f:compare_strategies_damping}}
    \end{overpic}
    \caption{\label{f:compare_strategies} Comparison of a single reaching movement to the same target position under the conditions FSVD, VSFD and VSVD. The figure shows the \begin{enumerate*}[label=(\alph*)]
        \item\label{f:compare_strategies_position} joint position,
        \item\label{f:compare_strategies_motor1} motor 1,
        \item\label{f:compare_strategies_motor2} motor 2, and
        \item\label{f:compare_strategies_damping} damping command
    \end{enumerate*}
    against time.}
\end{figure}

To examine this issue, the trajectories for reaching to a typical target position under the experimental conditions FSVD, VSFD and VSVD is plotted in \fref{compare_strategies}. Looking at \fref{compare_strategies}\ref{f:compare_strategies_position}, the accuracy is relatively good for each of the conditions (with some small overshoot in the FSVD condition). However, to achieve this accuracy, the FSVD controller, being unable to modulate speed by any other means, modulates the damping over its full range, maintaining a high damping constant during braking ($0.2<t<0.5\,s$), see \fref{compare_strategies}\ref{f:compare_strategies_damping}. As the rate of energy regeneration has its maximum at $u_3=0.5$, the more time the damping is held away from its medium value, the lower the total accumulated energy will be (see \sref{hybrid-damping}). In contrast, the damping command of VSVD remains low during the acceleration phase to decrease the loss and required input power, but then maintains the damping command close to the maximum regeneration damping level for the remainder of the movement. This is thanks to its ability to modulate the joint stiffness and thereby effect braking by an alternative means (see \fref{compare_strategies}\ref{f:compare_strategies_motor1} and \fref{compare_strategies}\ref{f:compare_strategies_motor2}). The variable stiffness therefore provides some flexibility in control to prevent the \emph{over-exploitation of variable damping} and the associated loss of regeneration.

\section{Conclusions}\label{s:conclusion}\noindent
This paper proposes an extension to variable damping module design for \VIAs\ based on the motor braking effect. In contrast to previous, pure dynamic braking designs, the proposed approach provides a solution for realising controllable damping, which enables the \VIAs\ to regenerate dissipated energy from bidirectional movement to charge a unidirectional electric storage element. Furthermore, it overcomes the drawback of a reduction in the maximum damping effect found in pure regenerative braking schemes. 

The control input for this damping module simply varies from $0$ to $1$, representing a proportional percentage of the maximum damping. As the power regeneration has a non-monotonic relation with the control input and damping coefficient (as verified by experiment), the balancing between damping allocation and energy regeneration needs to be treated with care. However, application of the hybrid damping module to \VIAs\ verified by experiments, 
shows that the actuation redundancy is solved by optimal control successfully to achieve fast smooth movement while still enabling power regeneration.

To investigate the use of variable regenerative damping for long-term operation, a stochastic consecutive reaching task was designed to examine the movement performance and energy efficiency. The experimental study shows that exploiting variable stiffness and variable damping is desired, in such a way that there is more flexibility to prevent over-exploitation of variable damping and loss of regeneration capability.

In future work, further prototyping of the power electronics and mechanical elements will be investigated to improve the transmission efficiency of the regenerative damping system. 
It will be tested on other more advanced VSA mechanisms.
The tasks considered in this paper are movements in free space without external perturbations. It is planned to investigate the use of the damping module for more complex long-term behaviours in presence of external perturbations or unpredicted environments, such as pick-and-place different objects with unknown weights, long-distance locomotion using a bipedal platform.
Furthermore, model learning techniques will be employed to produce accurate prediction of energy cost and regeneration for optimisation towards long-term energy efficiency.
Such information can be potentially used for high-level planning and human-guided learning control.
\bibliography{mybib170301,IEEEabrv}
\bibliographystyle{myIEEEtran}


\end{document}